\documentclass[sigconf]{acmart}
\usepackage{enumerate}
\usepackage{amsmath}

\usepackage{amssymb}
\usepackage{graphicx} 
\usepackage{subfigure}
\usepackage{multirow}

\usepackage{stfloats}
\usepackage{cuted}
\usepackage{capt-of}
\usepackage{marvosym}
\usepackage[noend]{algpseudocode}
\usepackage{algorithmicx,algorithm}
\usepackage{xcolor}

\newcommand{\zb}[1]{\textcolor{black}{#1}}

\usepackage{bbding}

\usepackage{fancyhdr}
\usepackage{hyperref}
\usepackage{color}
\AtBeginDocument{%
  \providecommand\BibTeX{{%
    \normalfont B\kern-0.5em{\scshape i\kern-0.25em b}\kern-0.8em\TeX}}}

\acmConference[MM '22] {Proceedings of the 30th ACM Int'l Conference on Multimedia}{October 10--14, 2022}{Lisbon, Portugal.}
\acmBooktitle{Proceedings of the 30th ACM Int'l Conference on Multimedia (MM '22), Oct. 10--14, 2022, Lisbon, Portugal.}

\acmSubmissionID{870}

\copyrightyear{2022}
\acmYear{2022}
\setcopyright{acmcopyright}\acmConference[MM '22]{Proceedings of the 30th ACM International Conference on Multimedia}{October 10--14, 2022}{Lisbon, Portugal}
\acmBooktitle{Proceedings of the 30th ACM International Conference on Multimedia (MM '22), October 10--14, 2022, Lisbon, Portugal}
\acmPrice{15.00}
\acmDOI{10.1145/3474085.3475532}
\acmISBN{978-1-4503-8651-7/21/10}

\begin{document}

\title{Learning Cross-Image Object Semantic Relation in Transformer for Few-Shot Fine-Grained Image Classification}

\author{Bo Zhang$^{1,4 \ast}$, Jiakang Yuan$^{1 \ast}$, Baopu Li$^3$, Tao Chen$^{1 \dagger}$, Jiayuan Fan$^2$, Botian Shi$^4$\\}
\affiliation{%
  \institution{$^1$School of Information Science and Technology, Fudan University, Shanghai, China \\
              $^2$Academy for Engineering and Technology, Fudan University, Shanghai, China \\
              $^3$Oracle Health and AI, Redwood City, USA \\
              $^4$Shanghai AI Laboratory, Shanghai, China \\
              }
  \country{\{bo.zhangzx, friskit.china\}@gmail.com, baopu.li@oracle.com, \{jkyuan18, eetchen, jyfan\}@fudan.edu.cn} 
  }

\thanks{$^{\ast}$ Bo~Zhang and Jiakang~Yuan are with equal contribution.\\
$^{\dagger}$ Tao~Chen is the corresponding author.}


\begin{abstract}
Few-shot fine-grained learning aims to classify a query image into one of a set of support categories with fine-grained differences.
Although learning different objects' local differences via Deep Neural Networks has achieved success, how to exploit the query-support cross-image object semantic relations in Transformer-based architecture remains under-explored in the few-shot fine-grained scenario. In this work, we propose a Transformer-based double-helix model, namely HelixFormer, to achieve the cross-image object semantic relation mining in a bidirectional and symmetrical manner. The HelixFormer consists of two steps: 1) Relation Mining Process (RMP) across different branches, and 2) Representation Enhancement Process (REP) within each individual branch. By the designed RMP, each branch can extract fine-grained object-level Cross-image Semantic Relation Maps (CSRMs) using information from the other branch, ensuring better cross-image interaction in semantically related local object regions. Further, with the aid of CSRMs, the developed REP can strengthen the extracted features for those discovered semantically-related local regions in each branch, boosting the model's ability to distinguish subtle feature differences of fine-grained objects. Extensive experiments conducted on five public fine-grained benchmarks demonstrate that HelixFormer can effectively enhance the cross-image object semantic relation matching for recognizing fine-grained objects, achieving much better performance over most state-of-the-art methods under 1-shot and 5-shot scenarios. Our code is available at: \textit{\textcolor{teal}{\url{https://github.com/JiakangYuan/HelixFormer}}}.
\end{abstract}




\maketitle

\section{Introduction}
\label{sec1}

The success of Deep Neural Networks (DNNs) \cite{He2016Deep, wen2021zigan, hu2018squeeze}, \cite{liu2016ssd, 9546634} largely owes to the completeness of training data, which means that the collected data should be carefully annotated, low level noised, and sufficiently large. But in many real scenarios that need professional labeling knowledge, such as fine-grained image classification for different kinds of birds, it is generally hard to access a large number of label-rich training samples.

One solution to alleviate the over-dependence of DNNs on label-rich training data is Few-Shot Learning (FSL)~\cite{sun2019meta,chen2019closer,ryu2020metaperturb,yang2020restoring,baik2020meta,yang2021free,liu2020negative}, which aims to mine the transferable meta-knowledge from the base classes so that DNNs can utilize such knowledge to easily recognize new classes given only a few training examples. However, the main challenge for the FSL \cite{sun2019meta,chen2019closer,ryu2020metaperturb,yang2020restoring,baik2020meta,yang2021free} is that learning to classify from few examples having limited representative capability inevitably brings the overfitting issue. Therefore, researchers mainly focus on leveraging meta-learning technology~\cite{vinyals2016matching, snell2017prototypical,ryu2020metaperturb, sung2018learning, hao2019collect, wu2019parn} to deal with the FSL problem. However, the above-mentioned FSL methods focus to classify coarse-grained generic object categories, which are less suitable to address the few-shot fine-grained classification task~\cite{wu2021object, zhu2020multi, wertheimer2021few, li2020bsnet}, that requires to emphasize the local feature variations or subtle feature differences.

\begin{figure}[t]
\vspace{-6pt}
\setlength{\abovecaptionskip}{0.1cm}
\setlength{\belowcaptionskip}{-0.2cm}
\centering
\includegraphics[height=4.5cm, width=8.3cm]{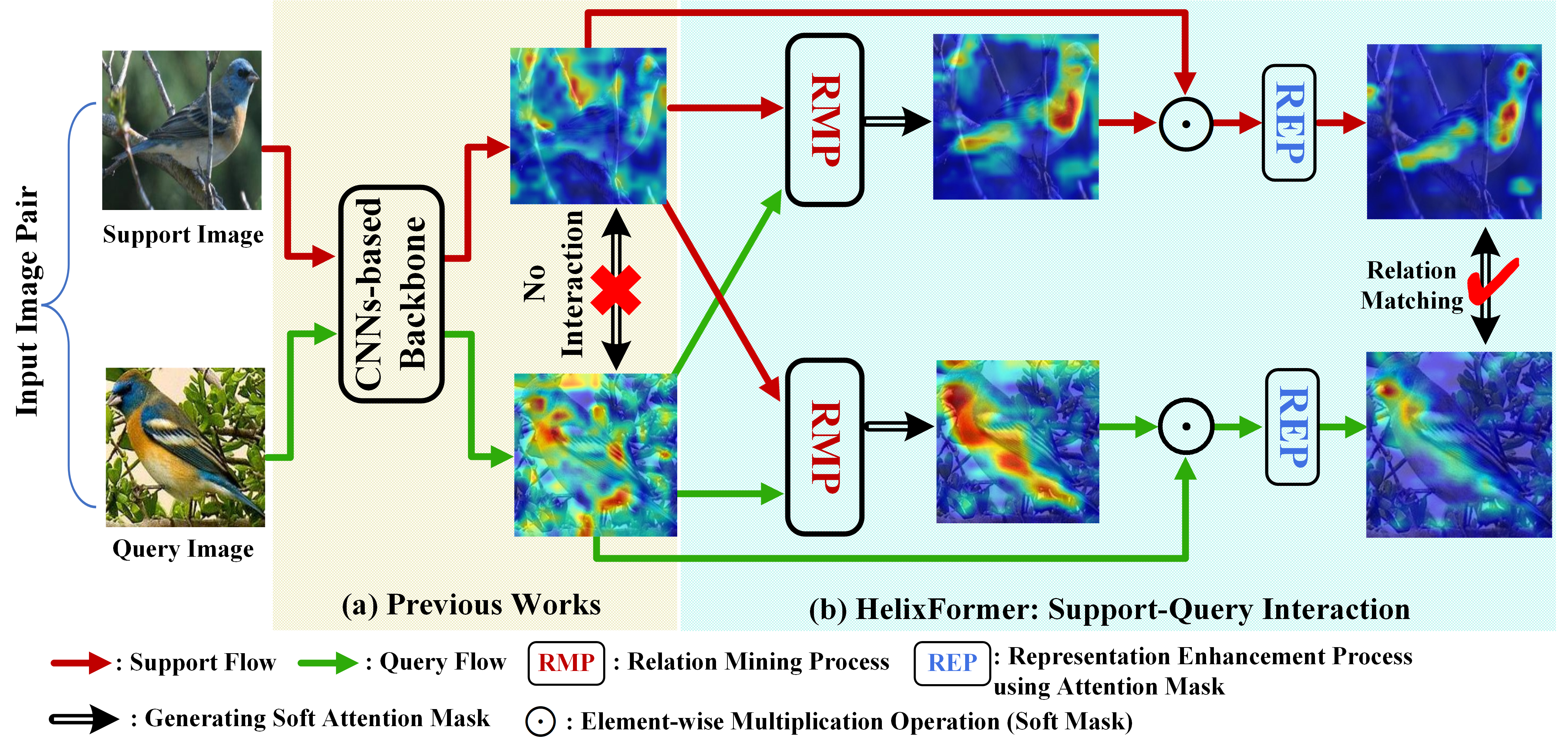}
\caption{(a) Previous fine-grained works attempt to learn discriminative local image parts but have no interaction between support-query images, which may cause relation matching confusion between different fine-grained objects and yield misclassification. (b) In contrast, HelixFormer consisting of RMP and REP modules (details will be given in Sec. \ref{sec3}) can mine image pair-level semantically related parts, \textit{e.g.} birds' wings, and further learn how to distinguish these subtle feature differences to avoid misclassification.}
\label{fig1}
\end{figure}

Inspired by the meta-learning success for generic object classification, some researchers~\cite{tang2020revisiting, wang2021few, li2020bsnet, zhu2020multi, wu2021object, dong2020learning, huang2020low, li2019revisiting, li2019distribution, garcia2017few, wertheimer2021few, tang2020blockmix, tang2022learning, huang2021toan} start to extend the study of FSL from generic object classes to fine-grained classes, where the main challenge is that fine-grained recognition is more dependent on \textbf{mining spatially local discriminative parts} of an input image, rather than global features extracted by generic meta-learning models such as prototypical networks~\cite{snell2017prototypical}. As illustrated in Fig. \ref{fig1}(a), many few-shot fine-grained works mine discriminative parts of the whole image based on the attention mechanism~\cite{zhu2020multi}, feature map reconstruction~\cite{wertheimer2021few}, and feature-level spatial alignment~\cite{wu2021object}. However, these methods fail to leverage cross-image object semantic relations between the training examples (denoting the support images) and test examples (denoting the query images).

As a matter of fact, \textit{to recognize a novel class's samples, humans tend to compare the local semantic parts' differences between the ever-seen and newly-seen objects, so that the subtle feature differences can be identified using only a few examples.} Thus, benefiting from such a fact, we are encouraged to model the cross-image object semantic relation to find the discriminative local regions in the Transformer, as illustrated in Fig. \ref{fig1}(b). We further give an in-depth analysis on how to design such a cross-attention mechanism under both few-shot and fine-grained scenarios via the following two aspects.

Firstly, under the FSL setting, we consider that the cross-image object semantic relation calculated from the support to query image, should be consistent with that calculated from the query to support image, namely, \textbf{cross-image object semantic relation consistency}, which motivates us to use a symmetrical structure to model the cross-image object semantic relations. Moreover, similar to humans that need to distinguish from multiple relations between a target object and its various neighbors to determine its belonged category, cross-image object semantic relation should be modeled as \zb{a parts-to-parts (many-to-many) matching process, that involves several attention heads of a Transformer running in parallel with each head focusing on a certain part.}

Secondly, such cross-image semantic relation learning should not only consider the few-shot scenario, but also consider the fine-grained setting which requires to strengthen subtle local features' discriminability. Different fine-grained objects in real world may present arbitrary poses, scales and appearances, causing different similarity in both global appearance and various local regions, hurting the classification performance. This challenge may be alleviated if the learned cross-image semantic relation can serve to emphasize those discriminative local features by integrating into a well-performed fine-grained baseline model, \zb{and enhance the representation learning process for those discovered semantically-similar regions.}

In view of the above, we propose a Transformer-based double-helix model, namely HelixFormer to solve the few-shot fine-grained image classification task. Benefiting from the self-attention mechanism of Transformer, the HelixFormer exploits the multi-head key-query-value outputs to do interaction between different images, and predicts their object-level semantic relation. HelixFormer is mainly composed of a cross-image object Relation Mining Process (RMP) across the support-query branches, and a feature Representation Enhancement Process (REP) within each single branch. Specifically, we feed the input image pairs into the support and query branches respectively. First, by treating the input feature patches (extracted by the Convolution Neural Networks (CNNs) based backbone such as Conv-4~\cite{sung2018learning}) as tokens and considering the cross-branch token information interaction, RMP produces \textbf{Cross-image Semantic Relation Maps (CSRMs)} for each branch in a bidirectional and symmetrical manner, which ensures the cross-image relation consistency. Meanwhile, we formulate the multi-head cross-attention mechanism as modeling many-to-many inter-object semantic relations, where one object would interact with multiple other objects. Second, the above CSRMs encode image patch-level semantic relation between different fine-grained objects, and further help the subsequent REP \zb{to enhance the learning of the feature representations within either the support or query feature encoding branch, which boosts the baseline model's ability to distinguish subtle feature differences of fine-grained objects.}

The main contributions of this work can be summarized as follows:

\begin{enumerate}[1)]
\item We propose a novel HelixFormer architecture, leveraging on the cross-image object semantic relation learning at patch level, to perform the few-shot fine-grained learning. To our knowledge, this is the first work to introduce semantic relation mining in the Transformer model for few-shot fine-grained task.

\item To ensure the semantic relation consistency between a pair of support-query images in FSL, we design a double-helix RMP structure that can generate consistent patch-level CSRMs in different branches. Furthermore, with the aid of CSRMs, we develop a REP to enhance the feature learning for those semantically-similar regions presenting subtle feature differences.

\item We have conducted extensive experiments on five few-shot fine-grained benchmarks. Both qualitative and quantitative results validate that the HelixFormer can effectively learn the cross-image object semantic relations, and further utilize such relations to enhance the model's generalization ability.
\end{enumerate}

\section{Related Works}
\label{sec2}

\noindent\textbf{Few-Shot Fine-Grained Learning (FSFGL).} Recent FSL works \cite{alet2019neural, yin2019meta, 9729102, ren2018learning, ye2022makes, lee2019learning, kirsch2019improving} can be roughly categorized into three types: 1) Optimization-based methods \cite{finn2017model,rusu2018meta} that focus on learning good initialization parameters in order to quickly adapt the few-shot model to novel classes; 2) Metric-based methods \cite{vinyals2016matching, snell2017prototypical, oreshkin2018tadam, sung2018learning} that aim to design a distance metric, so that the few-shot model can learn the semantic relation between different input images; 3) Data augmentation-based methods \cite{reed2017few, chen2019image, zhang2018metagan, hariharan2017low, wang2018low, li2020adversarial} that produce new samples to enlarge the training set for model training. Recently, inspired by the rapid development of meta-learning, researchers~\cite{tang2020revisiting, wang2021few, li2020bsnet, zhu2020multi, wu2021object, dong2020learning, huang2020low, li2019revisiting, li2019distribution, garcia2017few, wertheimer2021few} start to explore the generalization ability of FSL model on novel fine-grained sub-classes where only a few training examples are given. For example, a multi-attention meta-learning method~\cite{zhu2020multi} is proposed to learn diverse and informative parts of fine-grained images. Besides, the work~\cite{wertheimer2021few} tackles the FSFGL problem from the perspective of reconstructing the query image to learn a classifier. More recently, the work~\cite{wu2021object} tries to increase the fine-grained classification accuracy via long-shot-range spatial alignment between support and query features. Motivated by these works in the FSFGL community, we further extend the study of FSFGL to a Transformer-based structure, and investigate its effectiveness in strengthening the support-query relation matching process only given a few samples.

\noindent\textbf{Cross-attention Models.} In this part, we review existing cross-attention works \cite{chen2021crossvit, lin2021cat, wang2021crossformer, wei2020multi, Yang3690, ke2021prototypical, huang2019ccnet} and find that they are mainly based on the attention modeling of cross-scale features \cite{chen2021crossvit, lin2021cat, wang2021crossformer}, cross-modality relationships \cite{wei2020multi, Yang3690}, joint spatio-temporal information \cite{ke2021prototypical}, and inner-image multi-patches \cite{huang2019ccnet} to capture the intra-object relations and contextual information. Different from the above methods, our work aims to exploit the cross-image object semantic relations (\textit{i.e.} finding the discriminative local spatial regions between objects that are helpful for fine-grained recognition) to address the FSFGL issue. On the other hand, there are only two works \cite{hou2019cross, zhuang2020learning} employing cross-attention mechanism to perform the FSL task. Our work differs from the two works as follows: 1) We study a symmetrical Transformer-based structure, which fully considers the symmetry property between support and query images in FSL, and imposes a cross-image object semantic relation consistency between support-query and query-support matching process; 2) We develop a two-step relation matching process (a two-branch relation mining process and a representation enhancement process), which has the merit of improving the baseline model's ability to distinguish the subtle local feature differences.

\noindent\textbf{Transformer for Vision Tasks.} Vision Transformer (ViT) \cite{dosovitskiy2020image} shows promising performance on a variety of tasks including image classification \cite{dosovitskiy2020image}, object detection \cite{chu2021twins, wang2021pyramid}, segmentation \cite{chu2021twins, wang2021pyramid}, and pose estimation \cite{yuan2021hrformer}. The goal of ViT is to model long-range dependencies across different input sequence elements (or input tokens), by splitting an input image into a sequence of image patches with size of 16$\times$16 pixels. More recently, to reduce the computational cost, many researchers incorporate the multi-scale branch into Transformer, via inserting depth-wise convolutions into the self-attention \cite{yuan2021hrformer} or exploiting the multi-resolution parallel structure \cite{liu2021swin}. The above works attempt to model the global self-attention within an input image through the single-branch structure, while our HelixFormer tries to identify the local patch-level cross-attention between different input images. Besides, CrossTransformer \cite{doersch2020crosstransformers} and CrossViT \cite{chen2021crossvit} are recently developed Transformer-based dual-branch structures, where the CrossTransformer utilizes the dual-branch structure to achieve a coarse-grained spatial correspondence, and CrossViT feeds the image patches of different sizes into two separate branches to extract multi-scale information. We would like to emphasize that, compared with the above dual-branch network structure, our HelixFormer is actually a double-helix dual-branch structure, which means that the cross-object semantic relations from the support to query branch and vice versa are symmetric and complementary, ensuring the semantic consistency assumption of relation pairs.

\section{The Proposed Method}
\label{sec3}

The overall framework of the proposed HelixFormer is illustrated in Fig. \ref{fig2}. For easy understanding, we first give the problem formulation and the episodic training strategy for Few-Shot Learning (FSL) task. Then we introduce the proposed HelixFormer and discuss its superiority over several variants of existing cross-attention models. Finally, we give the overall objectives and cross-attention learning strategy of our model.

\subsection{Preliminaries}

\noindent\textbf{FSL Setting.} Given a set of base classes {\small $D_{base}$} and a CNN-based feature embedding network (or backbone) $F$, the purpose of few-shot learning is to learn a task-agnostic $F$ on {\small $D_{base}$} via an episodic task sampling strategy, so that the $F$ can be generalized to novel classes {\small $D_{novel}$}, where {\small $D_{base} \cap D_{novel} = \emptyset$}.
For a typical FSL setting, each episodic task represents an $N$-way $K$-shot classification task, where both support set $S$ and query set $Q$ are sampled from the same $N$ classes. During the meta-training stage, each episodic task is sampled from the base classes {\small $D_{base}$}.

\begin{figure*}
\vspace{-6pt}
\setlength{\abovecaptionskip}{0.1cm}
\setlength{\belowcaptionskip}{-0.5cm}
\centering
\includegraphics[height=6.4cm]{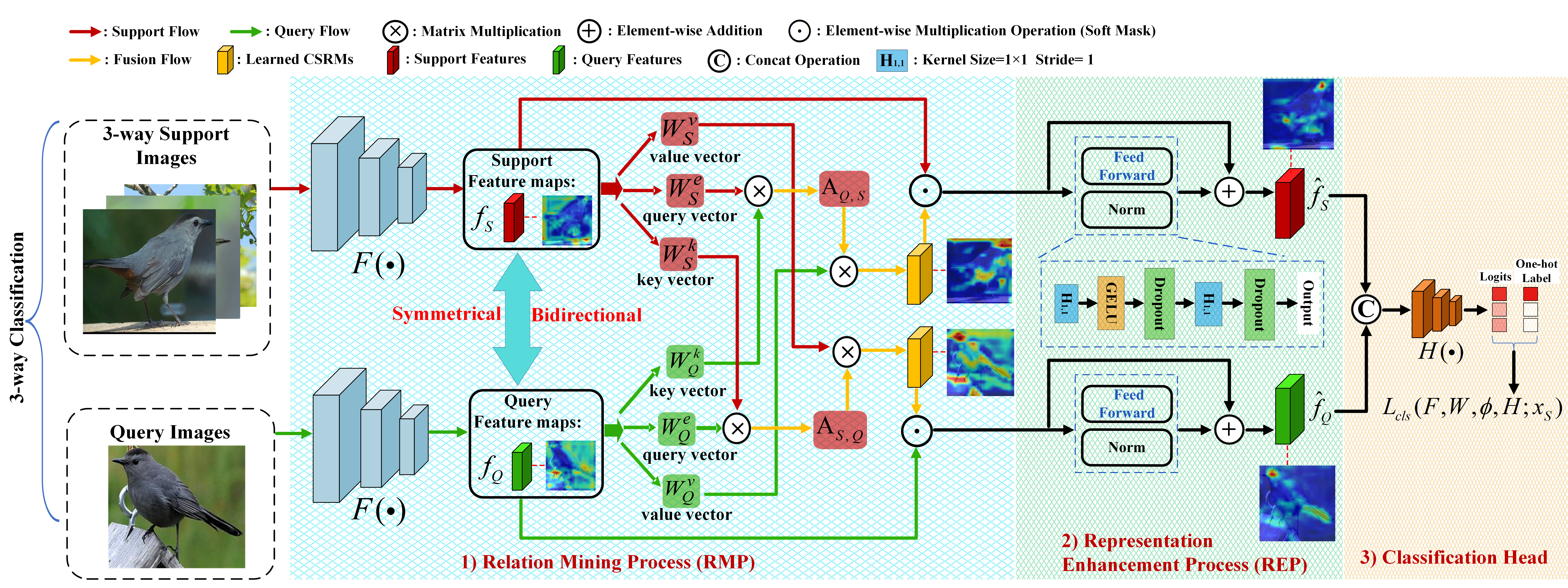}
\caption{The overview of the proposed HelixFormer, which takes a pair of support-query features {\small $(f_S, f_Q)$} extracted by a CNN backbone as its input, and outputs a pair of {\small $(\hat{f}_S, \hat{f}_Q)$} for doing the subsequent object classification. Note that for simplicity, we omit the multi-head attention in this figure.}
\label{fig2}
\end{figure*}

\noindent\textbf{Two-branch Baseline Introduction.} Inspired by the two-branch network structure to learn semantic representations in the relation network (RN)~\cite{sung2018learning}, we employ the RN as our baseline model. As illustrated in Fig. \ref{fig2}, given an input image pair $(x_S, x_Q)$, the RN first produces the high-level semantic feature pairs $(f_S, f_Q)$ via a convolution-based backbone $F$. Then, a classification network $H$ is used to predict whether the query image $x_Q$ has the same class label with the $n$-th class support image $x_{n,S}$. Thus, the loss function of our baseline model $L_{cls}(F, H;x_S)$ can be formulated as follows:

\begin{small}
\begin{equation}
\begin{aligned}
\label{eq1}
L_{cls}(F,H;x_S) = &\sum\nolimits_{(x_Q,y_Q)\in Q} log \ P(y_Q=n|x_Q;x_S) \\& = \frac{exp(H([F(x_{n,S}), F(x_Q)]))}{\sum_{n^{\prime} \in N} exp(H([F(x_{n^{\prime},S)}, F(x_Q)]))}
\end{aligned}
\end{equation}
\end{small}

\noindent where $[F(x_{n,S}), F(x_Q)]$ denotes the concat operation between $F(x_{n,S})$ and $F(x_Q)$, and $y_Q$ is the label of query image. During the meta-test or model evaluation stage, each episodic task is sampled from the novel classes {\small $D_{novel}$}, given the prototype of $K$ labeled support images of the $n$-class $x_{n,S}$. Note that the label $y_Q$ is available for model training on {\small $D_{base}$}, but it can only be used for model evaluation on {\small $D_{novel}$}.

\subsection{HelixFormer}
\label{sec3}

The purpose of this work is to capture the cross-image object semantic relations for improving the generalization ability of fine-grained model. By means of the multi-head attention module in Transformer, a many-to-many matching process of semantically related regions can be established. \textbf{Note that} \textit{just a single HelixFormer} is sufficient in capturing the cross-attention, and please refer to our supplementary material for the study of stacking HelixFormer.

\noindent\textbf{Bidirectional Relation Mining Process.} Given a pair of images $(x_S, x_Q)$ sampled from the $S$ and $Q$ respectively, the backbone $F$ is first used to extract a pair of high-level features $(f_S, f_Q)$ where $f_S = F(x_S) \in \mathbb{R}^{C\times H\times W}$, and $C$ denotes the channel number, $H$ and $W$ are the height and width of the features, respectively. Although the feature pairs $(f_S, f_Q)$ contain rich semantics, they lack interaction from each other, and do not consider the cross-object semantic relations between support-query images.

To fully encode the relations between the two branches, we treat the feature maps $f \in \mathbb{R}^{C\times H\times W}$ as a sequence of $HW$ tokens, with each token having $C$ channels, which can be formulated as $f = [f^1, f^2, ..., f^{HW}]$, where $f^i \in \mathbb{R}^{C}$.

In detail, given the token embedding with weight parameters $W_S^e$, $W_S^k$, $W_S^v$ for support branch $S$, and parameters $W_Q^e$, $W_Q^k$, $W_Q^v$ for query branch $Q$, the query vector $e^i$, key vector $k^i$, and value vector $v^i$ can be calculated as follows:

\begin{small}
\begin{equation}
\label{eq2.0}
\left\{  \begin{array}{l}
e_S^i = W_S^e \ f_S^i\\
k_S^i = W_S^k \ f_S^i\\
v_S^i = W_S^v \ f_S^i
\end{array} \right.\quad \quad \left\{ \begin{array}{l}
e_Q^i = W_Q^e \ f_Q^i\\
k_Q^i = W_Q^k \ f_Q^i\\
v_Q^i = W_Q^v \ f_Q^i
\end{array}  \right.
\end{equation}
\end{small}

\noindent where for avoiding the confusion of symbol definition, $e^i$ denotes the query vector in the RMP, and $Q$ represents the query branch for the whole pipeline. Besides, according to the ablation studies in Sec. \ref{sec4.4}, HelixFormer employs the convolution-based token embedding and removes the position embedding, since local spatial information has been encoded in the convolution feature maps. Further, we achieve the RMP by a symmetrical cross-attention from two directions: 1) We obtain support branch-related features using the value vector $v_Q^i$ from another branch $Q$, formulated as Q$\rightarrow$S; 2) We also obtain query branch-related features using the value vector $v_S^i$ of the support branch $S$, formulated as S$\rightarrow$Q.

\textbf{For Q$\rightarrow$S direction}, let $\mathbf{A}_{Q,S} \in \mathbb{R}^{HW \times HW}$ denote the matrix of attention scores obtained via the matrix multiplication as follows:

\begin{equation}
\label{eq2}
\mathbf{A}_{Q,S} = K_Q \ E_S^\mathrm{T}
\end{equation}

\noindent where $K_Q = [k_Q^1, ..., k_Q^{HW}] \in \mathbb{R}^{HW \times C}$ and $E_S = [e_S^1, ..., e_S^{HW}] \in \mathbb{R}^{HW \times C}$, which can be obtained using Eq. \ref{eq2.0}. Note that the designed token embedding way does not change the channel number for each input token $f^i \in \mathbb{R}^C$. Moreover, to perform normalization for attention scores in Transformer and find the semantically related regions according to clues from another branch $Q$, a softmax layer with a scaling factor is employed as follows:

\begin{small}
\begin{equation}
\label{eq3}
R_{Q,S} = Softmax(\mathbf{A}_{Q,S} / \sqrt{C}) \ V_Q
\end{equation}
\end{small}

\noindent where $V_Q = [v_Q^1, v_Q^2, ..., v_Q^{HW}] \in \mathbb{R}^{HW \times C}$, and $R_{Q,S} \in \mathbb{R}^{HW \times C}$ represents the Cross-image Semantic Relation Maps (CSRMs), encoding the patch-level semantic relations from query to support branch. Then, the CSRMs are reshaped to the same dimension as the backbone features $f_S \in \mathbb{R}^{C \times H \times W}$, in order to enhance the semantically similar backbone features in the REP operation.

\textbf{For S$\rightarrow$Q direction}, the CSRMs $R_{S,Q} \in \mathbb{R}^{HW \times C}$ also can be easily obtained by performing a symmetrical process described in Eqs. \ref{eq2} and \ref{eq3}, which can be written as follows:

\vspace{-0.10cm}
\begin{small}
\begin{equation}
\begin{array}{l}
\mathbf{A}_{S,Q} = K_S \ E_Q^\mathrm{T}, \\
R_{S,Q} = Softmax(\mathbf{A}_{S,Q} / \sqrt{C}) \ V_S.
\end{array}
\end{equation}
\end{small}

\noindent\textbf{Representation Enhancement Process.} \zb{Based on the above RMP, bidirectional semantic relations of support-to-query features and query-to-support features have been symmetrically encoded into the matrix of attention scores, from both directions $\mathbf{A}_{Q,S}$ and $\mathbf{A}_{S,Q}$, so that these cross-image object semantically-similar parts can be first found}. In this part, we design a REP that can further guide the classification network to learn how to distinguish these semantically similar features obtained by the RMP.

Given the high-level features $(f_S, f_Q)$ learned from the CNNs-based backbone, and the CSRMs $(R_{Q,S}, R_{S,Q})$ calculated from the Q$\rightarrow$S and S$\rightarrow$Q, the REP can be formulated as follows:

\begin{small}
\begin{equation}
\label{eq5}
\left\{  \begin{array}{l}
\hat{f}_S = MLP(Norm(f_S \odot R_{Q,S})) \\
\hat{f}_Q = MLP(Norm(f_Q \odot R_{S,Q}))
\end{array} \right.
\end{equation}
\end{small}

\noindent where $\odot$ denotes an element-wise multiplication operation, and CSRMs are defined as a soft relation mask that can strengthen the features $f$ extracted by CNNs-based backbone. Besides, $MLP$ is the Feed Forward module as illustrated in Fig. \ref{fig2}, which allows the backbone to focus on the subtle feature differences of the predicted semantically similar regions. The experimental analyses of the REP are shown in Sec. \ref{sec4.4}. Overall, \zb{$\hat{f}_S$ and $\hat{f}_Q$ are defined as the output features of the REP, and then will be fed into the classification head as illustrated in Fig. \ref{fig2}.}



\begin{figure*}
\vspace{-6pt}
\setlength{\abovecaptionskip}{0.1cm}
\setlength{\belowcaptionskip}{-0.3cm}
\centering
\includegraphics[height=5.8cm]{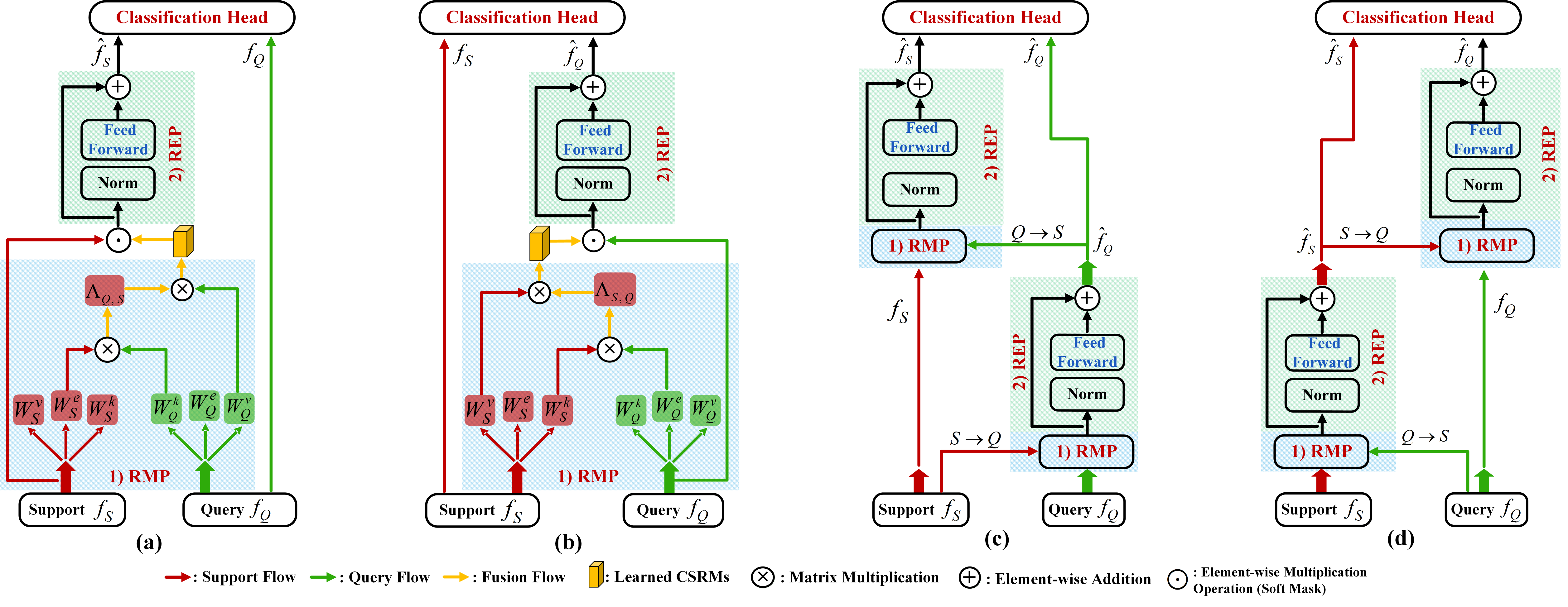}
\caption{Other Transformer-based cross-attention model designs. (a) Q$\rightarrow$S: Cross-attention from query to support; (b) S$\rightarrow$Q: Cross-attention from support to query; (c) S$\rightleftharpoons$Q: Bidirectional asymmetric cross-attention, which is a sequential stack of the above S$\rightarrow$Q and Q$\rightarrow$S variants; (d) Q$\rightleftharpoons$S: Bidirectional asymmetric cross-attention by stacking the Q$\rightarrow$S and S$\rightarrow$Q variants.}
\label{fig3}
\end{figure*}

\noindent\textbf{Differences among Transformer-based Cross-attention Variants.}
\label{sec3.3}
Given high-level feature pairs $(f_S, f_Q)$ extracted from CNN-based backbone $F$, there are many Transformer-based alternatives to model support-query patch-level relations of the extracted high-level features pairs $(f_S, f_Q)$. They can be categorized into three classes as follows.



\textbf{1)} \zb{Unidirectional cross-attention structure: As shown in Figs. \ref{fig3}(a) and \ref{fig3}(b), only the features from a single branch (support or query branch) are enhanced by means of cross-attention from another branch. For such a case, enhanced features $\hat{f}$ and original backbone features $f$ are used as the input of classification head. Such a unidirectional cross-attention way fails to achieve high classification accuracy, since this way only considers the semantic enhancement of only a single branch. }

\textbf{2)} \zb{Bidirectional but asymmetric structure: As illustrated in Figs. \ref{fig3}(c) and \ref{fig3}(d), S$\rightleftharpoons$Q (or Q$\rightleftharpoons$S) is a sequential stack of the above S$\rightarrow$Q and Q$\rightarrow$S variants (or Q$\rightarrow$S and S$\rightarrow$Q variants), where both the query features and support features are enhanced. But this approach does not consider support-query and query-support feature matching processes in parallel, which is detrimental to the cross-image object relation consistency assumption.}

\textbf{3)} \zb{Bidirectional and symmetrical structure: S$\leftrightarrow$Q refers to HelixFormer, which is shown in Fig.2. Compared with the above structures, HelixFormer is a more general form of cross-attention models, and thus has much less inductive bias. For unidirectional or bidirectional asymmetric structure, a kind of \textit{uneven/biased} cross-attention learning between support and query patch-level features is injected into the whole network. But for HelixFormer, the learned cross-image patch-level attention relations are symmetric and complementary. The detailed experimental and visual analyses are shown in Sec. \ref{sec4.4}}

\vspace{-0.20cm}
\subsection{Overall Objectives and Cross-attention Learning Strategy}

\noindent\textbf{Overall Objectives.} For finding a many-to-many semantic relation matching between the support and query branches, we utilize the multi-head attention mechanism, which consists of multiple attention layers with different token embedding parameters. The overall loss function of the $n$-th class on base classes {\small $D_{base}$} of HelixFormer can be written as follows:

\begin{small}
\begin{equation}
\begin{aligned}
\label{eq6}
L_{cls}(F, W, \phi, H; x_S) = &\sum\nolimits_{(x_Q,y_Q)\in Q} log \ P(y_Q=n|x_Q; x_S) \\& = \frac{exp(H([\hat{f}_{n,S},\hat{f}_Q]))}{\sum_{n^{\prime} \in N} exp(H([\hat{f}_{n^{\prime},S}, \hat{f}_Q]))}
\end{aligned}
\end{equation}
\end{small}

\noindent where $W$ and $\phi$ denote learnable parameters in RMP and REP, respectively.

\begin{table*}[]
\centering
\setlength{\tabcolsep}{1.25mm}{
\begin{tabular}{c|c|c|c|c|c|c|c|c}
\hline
\multirow{2}{*}{Method} & \multirow{2}{*}{Setting} & \multirow{2}{*}{Backbone} & \multicolumn{2}{c}{Stanford Dogs}         & \multicolumn{2}{c}{Stanford Cars}         & \multicolumn{2}{c}{NABirds}               \\
                        &            &               & 1-shot              & 5-shot              & 1-shot              & 5-shot              & 1-shot              & 5-shot              \\ \hline
RelationNet~(CVPR-18)~\cite{sung2018learning}   & In.          & Conv-4                   & \, 43.29±0.46$^\diamond$          & \, 55.15±0.39$^\diamond$          & \, 47.79±0.49$^\diamond$          & \, 60.60±0.41$^\diamond$          &  64.34±0.81*          & 77.52±0.60*          \\
GNN$^\dagger$~(ICLR-18)~\cite{garcia2017few}        & In.          & Conv-4                   & 46.98±0.98        & 62.27±0.95        & 55.85±0.97        & 71.25±0.89        & -                   & -                   \\
CovaMNet~(AAAI-19)~\cite{li2019distribution}    & In.   & Conv-4                   & 49.10±0.76          & 63.04±0.65          & 56.65±0.86          & 71.33±0.62          & 60.03±0.98*          & 75.63±0.79*          \\
DN4~(CVPR-19)~\cite{li2019revisiting}         & In.    & Conv-4                   & 45.73±0.76          & 66.33±0.66          & 61.51±0.85          & 89.60±0.44 & 51.81±0.91*          & 83.38±0.60*          \\
LRPABN~(TMM-20)~\cite{huang2020low}       & In.   & Conv-4                   & 45.72±0.75          & 60.94±0.66          & 60.28±0.76          & 73.29±0.58          & 67.73±0.81*          & 81.62±0.58*          \\
MattML~(IJCAI-20)~\cite{zhu2020multi}     & In.     & Conv-4                   & 54.84±0.53          & 71.34±0.38          & 66.11±0.54          & 82.80±0.28          & -                   & -                   \\
ATL-Net~(IJCAI-20)~\cite{dong2020learning}       & In.       & Conv-4                   & 54.49±0.92          & 73.20±0.69          & 67.95±0.84          & 89.16±0.48          & -                   & -                   \\
FRN~(CVPR-21)~\cite{wertheimer2021few}       & In.       & Conv-4                   & 49.37±0.20 & 67.13±0.17 & 58.90±0.22 & 79.65±0.15          & - & - \\
LSC+SSM~(ACM MM-21)~\cite{wu2021object}       & In.       & Conv-4                   & 55.53±0.45 & 71.68±0.36 & 70.13±0.48 & 84.29±0.31          & 75.60±0.49 & 87.21±0.29 \\
Ours         & In.        & Conv-4                   & \textbf{59.81}±0.50 & \textbf{73.40}±0.36 & \textbf{75.46}±0.37 &  \textbf{89.68}±0.25        & \textbf{78.63}±0.48 & \textbf{90.06}±0.26 \\ \hline
LSC+SSM~(ACM MM-21)~\cite{wu2021object}       & In.       & ResNet-12                   & 64.15±0.49 & 78.28±0.32 & 77.03±0.46 & 88.85±0.46          & 83.76±0.44 & 92.61±0.23 \\
Ours      & In.            & ResNet-12                 & \textbf{65.92}±0.49 &  \textbf{80.65}±0.36 & \textbf{79.40}±0.43 & \textbf{92.26}±0.15          & \textbf{84.51}±0.41 & \textbf{93.11}±0.19 \\ \hline
\end{tabular}}
\caption{\label{tab1}5-way classification accuracy ($\%$) on the Stanford Dogs, Stanford Cars and NABirds datasets respectively, $^\diamond$, and * represent that the corresponding results are reported in~\cite{zhu2020multi}, and~\cite{huang2020low}, respectively. Other results are reported in their original papers. ``\;In.\;'' denotes the inductive few-shot learning.}
\vspace{-0.4cm}
\end{table*}

\begin{table}[]
\centering
\setlength{\tabcolsep}{0.8mm}{
\begin{tabular}{c|c|c|c}
\hline
\multirow{2}{*}{Method} & \multirow{2}{*}{Backbone} & \multicolumn{2}{c}{CUB}                   \\
                        &                           & 1-shot              & 5-shot              \\ \hline
FEAT~(CVPR-20)~\cite{ye2020few}                   & Conv-4                   & 68.87±0.22          & 82.90±0.15          \\
CTX~(NIPS-20)~\cite{doersch2020crosstransformers}                      & Conv-4                   & 69.64 & 87.31 \\
FRN~(CVPR-21)~\cite{wertheimer2021few}                    & Conv-4                   & 73.48 & 88.43 \\
LSC+SSM~(ACM MM-21)~\cite{wu2021object}                   & Conv-4                   &  73.07±0.46 & 86.24±0.29 \\
Ours                    & Conv-4                   & \textbf{79.34}±0.45 & \textbf{91.01}±0.24 \\ \hline
RelationNet*~(CVPR-18)~\cite{sung2018learning}             & ResNet-34                 & 66.20±0.99          & 82.30±0.58          \\
DeepEMD~(CVPR-20)~\cite{zhang2020deepemd}              & ResNet-12                 & 75.65±0.83          & 88.69±0.50          \\
ICI~(CVPR-20)~\cite{wang2020instance}                     & ResNet-12                 & 76.16               & 90.32               \\
CTX~(NIPS-20)~\cite{doersch2020crosstransformers}                      & ResNet-12                   & 78.47  & 90.90 \\
FRN (Baseline)~\cite{wertheimer2021few}                     & ResNet-12                   & 80.80±0.20 & - \\
FRN~(CVPR-21)~\cite{wertheimer2021few}                     & ResNet-12                   & \textbf{83.16} & \textbf{92.59} \\
LSC+SSM~(ACM MM-21)~\cite{wu2021object}                     & ResNet-12                   &  77.77±0.44 & 89.87±0.24 \\
Ours (Baseline)                & ResNet-12                 & 72.61±0.47 & 85.60±0.29  \\
Ours                    & ResNet-12                 & 81.66±0.30 & 91.83±0.17 \\ \hline
\end{tabular}}
\caption{\label{tab2}5-way classification accuracy ($\%$) on the CUB (using bounding-box cropped images). ``\;FRN (Baseline)\;'' represents the classification results achieved by \textbf{their baseline model}.}
\vspace{-0.4cm}
\end{table}

\begin{table}[]
\centering
\setlength{\tabcolsep}{1.5mm}{
\begin{tabular}{c|c|c|c}
\hline
\multirow{2}{*}{Method} & \multirow{2}{*}{Backbone} & \multicolumn{2}{c}{Aircraft}                   \\
                        &                           & 1-shot              & 5-shot              \\ \hline
ProtoNet~(NIPS-17)~\cite{snell2017prototypical}               & Conv-4                   & 47.72          & 69.42          \\
DSN~(CVPR-20)~\cite{simon2020adaptive}                   & Conv-4                   & 49.63          & 66.36          \\
CTX~(NIPS-20)~\cite{doersch2020crosstransformers}                    & Conv-4                   & 49.67 & 69.06 \\
FRN~(CVPR-21)~\cite{wertheimer2021few}                    & Conv-4                   &  53.20 & 71.17 \\
Ours                    & Conv-4                   & \textbf{70.37}±0.57 & \textbf{79.80}±0.42 \\ \hline
ProtoNet~(NIPS-17)~\cite{snell2017prototypical}               & ResNet-12                  & 66.57          & 82.37          \\
DSN~(CVPR-20)~\cite{simon2020adaptive}                    & ResNet-12                   & 68.16         & 81.85          \\
CTX~(NIPS-20)~\cite{doersch2020crosstransformers}                      & ResNet-12                   & 65.60 & 80.20 \\
FRN~(CVPR-21)~\cite{wertheimer2021few}                    & ResNet-12                   &  70.17  & \textbf{83.81}  \\
Ours                    & ResNet-12                 & \textbf{74.01}±0.54 & 83.11±0.41 \\ \hline
\end{tabular}}
\caption{\label{tab3}5-way classification accuracy ($\%$) on the Aircraft dataset.}
\vspace{-0.2cm}
\end{table}

\noindent\textbf{Cross-attention Learning Strategy.} To transfer the semantic relations from base classes {\small $D_{base}$} to novel classes {\small $D_{novel}$}, we employ a two-stage training strategy. Firstly, to ensure that the backbone $F$ has sufficient high-level semantic knowledge for subsequent cross-attention matching process, the $F$ is trained on base classes {\small $D_{base}$} by optimizing Eq. \ref{eq1}. Secondly, we insert the HelixFormer at the end of backbone $F$, and finetune the entire framework to compare the support and query images by optimizing Eq. \ref{eq6} on {\small $D_{base}$}.

\section{Experiments}
We evaluate the proposed method on five FSFGL benchmarks including Stanford Dogs, Stanford Cars, NABirds, CUB, and Aircraft. Additionally, we also perform cross-domain few-shot experiments to further show the transferability and adaptability of the HelixFormer. The following experiments are implemented by Pytorch, and all images are resized to 84$\times$84 pixels for fair comparison.

\vspace{-0.10cm}
\subsection{Dataset}
\label{sec4.1}
\vspace{-0.10cm}
\noindent\textbf{Stanford Dogs} \cite{khosla2011novel} contains a total of 20580 images and 120 sub-classes of dogs. Following~\cite{zhu2020multi}, we adopt 70, 20, 30 classes for meta-train, meta-validation and meta-test, respectively. \textbf{Stanford Cars} \cite{krause20133d} consists of 16,185 images from 196 sub-classes, and we adopt 130, 17, 49 classes for meta-train, meta-validation and meta-test, respectively. \textbf{NABirds} \cite{van2015building} provides 555 sub-classes of birds from North American. We use 350, 66, 139 categories for meta-train, meta-validation and meta-test, respectively, which is consistent with~\cite{huang2020low}. \textbf{CUB} \cite{wah2011caltech} has 11,788 bird images containing 200 classes. We follow the commonly used split way \cite{tang2020revisiting, wertheimer2021few}, which employs 100, 50, and 50 classes for meta-train, meta-validation and meta-test, respectively. \textbf{Aircraft} \cite{maji2013fine} includes 100 classes and 10000 aircraft images. Following the split way in \cite{wertheimer2021few}, we use 50, 25, and 25 classes for meta-train, meta-validation and meta-test, respectively.

\vspace{-0.25cm}
\subsection{Experimental Setup}
\label{sec4.2}
\vspace{-0.10cm}
\noindent\textbf{Stage One: Pre-training Backbone.} Following the setting in \cite{chen2020new, snell2017prototypical, rusu2018meta, ye2020few}, we insert a fully-connected layer at the end of the selected backbone such as Conv-4 or ResNet-12, and train the backbone on base classes {\small $D_{base}$}. In this stage, the backbone is trained from scratch using SGD optimizer with a batch size of 128, a momentum of 0.9, a weight decay of 0.0005, and an initial learning rate of 0.1. To keep consistent with the setting in \cite{wu2021object}, the learning rate decays at 85 and 170 epochs. We remove the fully-connected layer for performing the next meta-training stage.

\noindent\textbf{Stage Two: Meta-training HelixFormer.} In this stage, we first insert the proposed HelixFormer at the end of the pre-trained backbone, and then finetune the whole model to perform cross-attention for each input image pair. A learning rate of 0.001 is employed for all modules. SGD with a weight decay of 0.001 and Adam are used to optimize the backbone and HelixFormer, respectively. The whole training process lasts 130 epochs, and the learning rate decays to 0.0001 and 0.00001 at 70 and 110 epochs, respectively. \zb{The number of multi-head attention in the single-layer HelixFormer is set to $2$, and the corresponding experimental analysis is shown in Sec. \ref{sec4.4}}. For model evaluation, we report the results with 95$\%$ confidence intervals over 2000 test episodes, and the best model is chosen according to its accuracy on the validation set.

\vspace{-0.20cm}
\subsection{Experimental Results}
\label{sec4.3}
\vspace{-0.10cm}

\noindent\textbf{Few-shot Fine-grained Image Classification Results.}
It is generally recognized that spatially local feature relations across different fine-grained objects are particularly important for fine-grained classification. Thus, we first validate the effectiveness of HelixFormer on a wide range of fine-grained benchmark datasets, which is shown in Tables \ref{tab1}-\ref{tab3}.

First, Table \ref{tab1} reports the classification accuracies on the Standard Dogs, Standard Cars, and NABirds. It can be seen from Table \ref{tab1} that the proposed HelixFormer can be applied on different backbones such as Conv-4 \cite{sung2018learning} and ResNet-12 \cite{chen2019closer}. Moreover, we compare the proposed HelixFormer with state-of-the-art general few-shot learning methods (including DN4 \cite{li2019revisiting} and FRN \cite{wertheimer2021few}, \textit{etc.}) and few-shot fine-grained image classification methods (including MattML \cite{zhu2020multi}, LSC+SSM \cite{wu2021object}, \textit{etc.}). FRN considers the few-shot learning as a feature reconstruction problem by reducing the reconstruction errors of images belonging to the same classes, while LSC+SSM attempts to align spatial distribution of feature maps between support and query images, both of which are state-of-the-art general and fine-grained few-shot classification methods, respectively. The experimental results show that the proposed HelixFormer outperforms these methods with a considerable margin, further demonstrating that learning the cross-attention using the proposed double-helix structure can boost the accuracy of fine-grained object recognition.

Furthermore, we also conduct experiments on two more challenging datasets (CUB and Aircraft), as shown in Table \ref{tab2} and \ref{tab3}. We also observe a consistent accuracy increase using our method. Besides, we would like to emphasize that the HelixFormer significantly boosts the accuracy of Baseline from $72.61\%$ to $81.66\%$ on CUB dataset and has been verified on \textit{\textbf{five} commonly used fine-grained datasets}, comprehensively showing the effectiveness of HelixFormer on fine-grained recognition.

\noindent\textbf{Few-shot Fine-grained Cross-domain Image Classification Results.}
Considering that the distribution differences between the training and test data often exist, we conduct few-shot fine-grained recognition experiments under cross-domain setting, to validate the effectiveness of the HelixFormer in alleviating the impact of the domain differences, and the results are reported in Table \ref{tab4}.

We carry out the cross-domain adaptation from generic bird categories (widely collected from Internet) to a particular country (America). It can be seen from Table \ref{tab4} that our method has higher accuracy than both the Baseline model and LSC+SSM model \cite{wu2021object}, demonstrating that HelixFormer also can improve the transferability and domain adaptability of the existing models.

\begin{table}[t]
\centering
\setlength{\tabcolsep}{0.8mm}{
\begin{tabular}{c|c|c|c}
\hline
\multirow{2}{*}{Method} & \multirow{2}{*}{Backbone} & \multicolumn{2}{c}{CUB $\to$ NABirds} \\
                        &                           & 1-shot                 & 5-shot                 \\ \hline
LSC+SSM~(Baseline)~\cite{wu2021object}    & ResNet-12                 & 45.70±0.45    & 63.84±0.40                        \\
LSC+SSM~(ACM MM-21)~\cite{wu2021object}    & ResNet-12                 & 48.50±0.48    & 66.35±0.41                        \\ \hline
Baseline (RN~\cite{sung2018learning})               & Conv-4                 & 43.55±0.45            & 55.53±0.42            \\
Ours                & Conv-4                 & \textbf{47.87}±0.47            & \textbf{61.47}±0.41             \\\hline
Baseline (RN~\cite{sung2018learning})                   & ResNet-12                 &  46.22±0.45    & 63.23±0.42                        \\
Ours                    & ResNet-12                 & \textbf{50.56}±0.48    &  \textbf{66.13}±0.41                        \\ \hline
\end{tabular}
\caption{\label{tab4} 5-way few-shot fine-grained classification results by adapting from the CUB-trained model to NABirds dataset using different backbones.}
}
\vspace{-0.6cm}
\end{table}

\begin{figure}
    \centering
    \includegraphics[height=4.8cm, width=6.0cm]{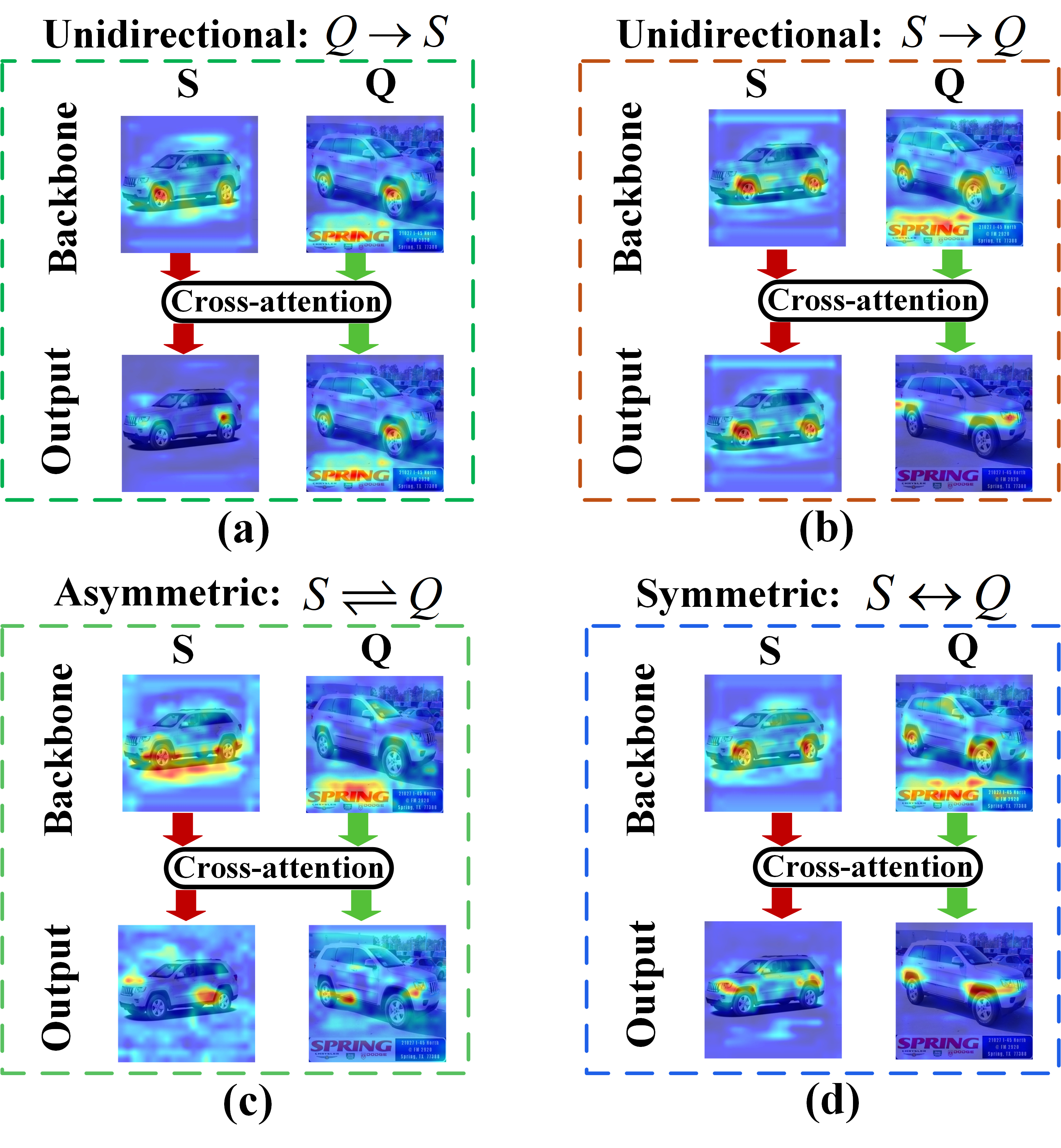}
    \vspace{-6pt}
 \caption{Visualization results of the backbone features and output features using different cross-attention model designs, respectively.}
 \label{fig4}
\end{figure}

\vspace{-0.10cm}
\subsection{Insight Analyses}
\label{sec4.4}
\vspace{-0.10cm}

\begin{table}[]
\centering
\setlength{\tabcolsep}{0.2mm}{
\begin{tabular}{c|c|c|c|c|c}
\hline
\multirow{2}{*}{Method} & \multirow{2}{*}{Token} & \multicolumn{2}{c}{CUB}   & \multicolumn{2}{c}{Stanford Cars}                \\
                        &                           & 1-shot              & 5-shot & 1-shot              & 5-shot              \\ \hline
B/L (RN~\cite{sung2018learning})        & -                   & 66.08±0.50          & 79.04±0.35    & 56.55±0.52          & 70.52±0.39        \\
RN with S$\leftrightarrow$Q          & Fc.                & 73.80±0.47 & 89.34±0.27     & 72.94±0.50 & 88.85±0.26 \\
RN with S$\leftrightarrow$Q          & Cv.                & \textbf{79.34}±0.45 & \textbf{91.01}±0.24    & \textbf{75.46}±0.37 & \textbf{89.68}±0.25 \\ \hline
\end{tabular}}
\caption{\label{tab5} The study of token embedding by employing fully-connected embedding (\textit{i.e.} Fc.) or convolutional embedding (\textit{i.e.} Cv.), respectively. B/L denotes the baseline model~\cite{sung2018learning}, and S$\leftrightarrow$Q denotes the proposed HelixFormer.}
\end{table}

\noindent\textbf{Results of using Fully-connected or Convolutional-based Token Embedding.}
The choice on feature embedding way (in a fully-connected or convolutional way) of the input tokens is essential to guarantee the good performance of the proposed HelixFormer. Table \ref{tab5} reports the results of using different embedding ways, showing that the accuracy using convolution projection outperforms that using fully-connected token embedding. The reason is that few-shot fine-grained recognition needs to capture more local detail features, which are exactly provided by the local convolution projection.

\begin{figure*}
\vspace{-6pt}
    \centering
    \includegraphics[height=4.0cm, width=17.3cm]{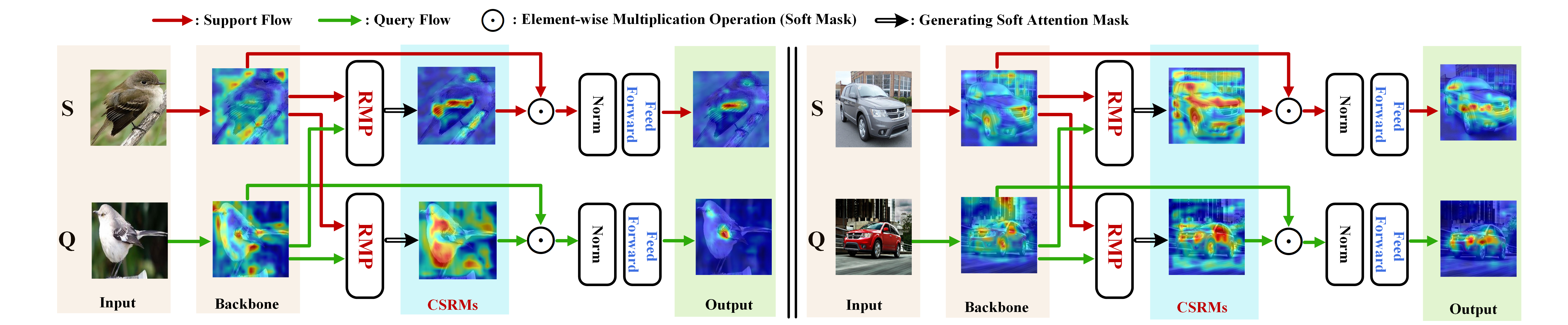}
 \vspace{-6pt}
 \caption{Visualization results of features extracted by \zb{backbone, the RMP, and the REP, respectively}. Due to the global feature variations (\textit{e.g.,} the color of an object) and local feature changes (\textit{e.g.,} headlight translation of a car and beak rotation of a bird), it is hard to find the cross-image patch-level matching of semantic features, as shown in the heatmaps from the backbone. By HelixFormer, the key cross-image object semantic relations, such as birds' wings or cars' headlights, can be effectively matched. \textit{Please refer to our supplementary material for more visualization results.}}
 \label{fig5}
\end{figure*}

\begin{table}[]
\centering
\setlength{\tabcolsep}{0.8mm}{
\begin{tabular}{c|c|c|c|c}
\hline
\multirow{2}{*}{Method}  & \multicolumn{2}{c}{CUB}   & \multicolumn{2}{c}{Stanford Cars}                \\
                        &     1-shot              & 5-shot & 1-shot              & 5-shot              \\ \hline
RN~\cite{sung2018learning}       & 66.08±0.50          & 79.04±0.35    & 56.55±0.52          & 70.52±0.39        \\
RN with Q$\rightarrow$S                   & 75.37±0.46   &  88.78±0.26         & 72.12±0.50  &  88.23±0.26        \\
RN with S$\rightarrow$Q         &     77.24±0.49      & 90.06±0.27 &  73.34±0.51  &     89.44±0.26      \\
RN with S$\leftrightarrow$Q            & \textbf{79.34}±0.45 & \textbf{91.01}±0.24     & \textbf{75.46}±0.37 & \textbf{89.68}±0.25 \\ \hline
\end{tabular}}
\caption{\label{tab6} Results using different cross-attention structures: the unidirectional cross-attention (including Q$\rightarrow$S and S$\rightarrow$Q) and the bidirectional structure. Q$\rightarrow$S denotes that support features are reconstructed only using query images, according to the semantic relations between support-query features, and vice versa. The definition of S$\leftrightarrow$Q follows Table~\ref{tab5}.}
\vspace{-0.2cm}
\end{table}

\begin{table}[]
\centering
\setlength{\tabcolsep}{0.38mm}{
\begin{tabular}{c|c|c|c|c|c}
\hline
\multirow{2}{*}{Method} & \multirow{2}{*}{SY?} & \multicolumn{2}{c}{CUB}   & \multicolumn{2}{c}{Stanford Cars}                \\
                        &                           & 1-shot              & 5-shot & 1-shot              & 5-shot              \\ \hline
RN~\cite{sung2018learning}           & -                   & 66.08±0.50          & 79.04±0.35    & 56.55±0.52          & 70.52±0.39        \\
RN with S$\rightleftharpoons$Q          & \scriptsize{\XSolid}              & 77.69±0.48 & 89.39±0.27    & 74.56±0.50 & \textbf{89.89}±0.22 \\
RN with Q$\rightleftharpoons$S          & \scriptsize{\XSolid}              & 77.46±0.47 & 89.86±0.25    & 74.42±0.50    & 88.63±0.24  \\
RN with S$\leftrightarrow$Q          & \Checkmark               & \textbf{79.34}±0.45 &   \textbf{91.01}±0.24   & \textbf{75.46}±0.37 & 89.68±0.25 \\ \hline
\end{tabular}}
\caption{\label{tab7} The study of bidirectional cross-attention using asymmetric or symmetric structure. SY is short for symmetric, and S$\rightleftharpoons$Q denotes the asymmetric but bidirectional structure and the definition of S$\leftrightarrow$Q follows Table~\ref{tab5}.}
\vspace{-12pt}
\end{table}

\begin{table}[]
\centering
\small
\setlength{\tabcolsep}{0.65mm}{
\begin{tabular}{c|c|c|c|c|c}
\hline
\multirow{2}{*}{Method} & \multirow{2}{*}{REP?} & \multicolumn{2}{c}{CUB}   & \multicolumn{2}{c}{Stanford Cars}                \\
                        &                           & 1-shot              & 5-shot & 1-shot              & 5-shot              \\ \hline
RN~\cite{sung2018learning}        & -                   & 66.08±0.50          & 79.04±0.35    & 56.55±0.52          & 70.52±0.39        \\
RN with S$\leftrightarrow$Q          & w/o                & 77.90±0.45   & 89.39±0.27     & 73.52±0.49  &  88.48±0.25  \\
RN with S$\leftrightarrow$Q          & with                & \textbf{79.34}±0.45 & \textbf{91.01}±0.24    & \textbf{75.46}±0.37 & \textbf{89.68}±0.25 \\ \hline
\end{tabular}}
\caption{\label{tab8} The results of removing the REP in HelixFormer.}
\vspace{-12pt}
\end{table}

\noindent\textbf{Unidirectional or Bidirectional RMP.}
In this part, we study the impact of using unidirectional or bidirectional cross-attention structure (as introduced in Fig. \ref{fig3} of Sec. \ref{sec3.3}) on the classification results. Experimental results in Table \ref{tab6} show that compared with the unidirectional structure (Q$\rightarrow$S or S$\rightarrow$Q), the bidirectional cross-attention structure (S$\leftrightarrow$Q) can well improve the generalization ability of few-shot learning models.

\noindent\textbf{Asymmetric and Symmetric RMP.} For bidirectional attention, we observe from Table \ref{tab7} that a symmetrical structure has an advantage in improving the model accuracy. We further visualize their heatmaps in Fig. \ref{fig4}, and find that the symmetrical structure captures relatively more accurate inter-object semantic relations.

\noindent\textbf{The Role of REP.} We further show the effectiveness of the REP in the proposed HelixFormer, by directly feeding the CSRMs pairs $(R_{Q,S}, R_{S,Q})$ as the input of the classification head. We observe the performance deterioration by comparing the last two rows in Table \ref{tab8}, showing the importance of learning to distinguish subtle feature differences by the REP.

\noindent\textbf{Multi-head Attention in HelixFormer.} Table \ref{tab9} shows the 5-way 1-shot classification accuracy by changing the number of multi-head attention using Conv-4 backbone.

\begin{table}[]
\centering
\setlength{\tabcolsep}{1.2mm}{
\begin{tabular}{c|c|c|c}
\hline
Method & \# Multi-head &  CUB     & Stanford Cars                  \\ \hline
RN with S$\leftrightarrow$Q       & 1  & 77.52±0.49  & 74.62±0.50   \\
RN with S$\leftrightarrow$Q       & 2  & \textbf{79.34}±0.45 &  \textbf{75.46}±0.37     \\
RN with S$\leftrightarrow$Q       & 4    & 78.68±0.47  &  73.62±0.48  \\ \hline
\end{tabular}}
\caption{\label{tab9} Results of changing the number of multi-head attention.}
\vspace{-0.2cm}
\end{table}

\begin{table}[]
\small
\centering
\setlength{\tabcolsep}{1.0mm}{
\begin{tabular}{c|c|c|c|c}
\hline
Method & Backbone & \#FLOPs. & \#Params.  & CUB                 \\ \hline
RN~\cite{sung2018learning}  & ResNet-12 & 2.48G & 9.1M & 72.61±0.47       \\
RN~\cite{sung2018learning}   & ResNet-50 (\textbf{Deeper}) & 3.69G & 24.6M & 69.00±0.52        \\
RN~\cite{sung2018learning}   & ResNet-101 (\textbf{Deeper}) & 5.98G & 43.2M & 68.71±0.54        \\
RN with S$\leftrightarrow$Q     & ResNet-12  & 2.53G   & 9.5M   & \textbf{81.66}±0.30   \\ \hline
\end{tabular}}
\caption{\label{tab10} 5-way 1-shot classification accuracy (\%) for relation network baseline~\cite{sung2018learning} with different backbones.}
\vspace{-0.6cm}
\end{table}

\noindent\textbf{Study of Parameter Size and Feature Visualization.}
In the few-shot learning community, models with smaller parameter sizes are usually adopted to avoid over-fitting for the limited number of training samples. In other words, increasing the number of model parameters may not improve its generalization performance. Table \ref{tab10} shows that the HelixFormer more effectively boosts the model accuracy, compared with other models with more parameters and FLOPs. Furthermore, we visualize the support and query features which are extracted from the backbone, the proposed RMP, and REP of HelixFormer respectively. The visualized results are illustrated in Fig. \ref{fig5}, and the heatmaps are obtained via a max-pooling operation along the channel dimension of feature maps to preserve the spatial information. Besides, we also visualize the support-query features using asymmetric or symmetric cross-attention structure in Fig. \ref{fig5}. These visualization results illustrate that the semantically similar regions between support and query images can be well matched via HelixFormer.

\section{Conclusion}
In this work, we proposed a Transformer-based double-helix cross-attention model, namely HelixFormer, which is composed of a Relation Mining Process (RMP) to discover the cross-image object semantic relation and produce patch-level cross-relation maps, and a Representation Enhancement Process (REP) to enhance the identified discriminative local features for final recognition. Experimental results demonstrate that such a bidirectional and symmetrical structure has the merit of ensuring the cross-object semantic relation consistency, improving the model generalization ability on few-shot fine-grained learning.

\clearpage
\section{Acknowledgement}
This work is supported by National Natural Science Foundation of China (No. 62071127 and No. 62101137), Zhejiang Lab Project (No. 2021KH0AB05).




{\small
\bibliographystyle{ACM-Reference-Format}
\bibliography{egbib}
}

\begin{appendix}
\clearpage
\section*{Outlines}

In the supplementary material, we provide more details and experimental results of the proposed HelixFormer, which includes two aspects:

\begin{enumerate}[1)]

\item The study of stacking HelixFormer at the end of CNN-based backbone;

\item Detailed experimental settings of training HelixFormer and more visualization results of HelixFormer.
\end{enumerate}


\section{Stacking HelixFormer}

\setcounter{table}{10}
\begin{table}[!h]
\label{tab11}
\centering
\setlength{\tabcolsep}{1.0mm}{
\begin{tabular}{cccc}
\hline
Method & \# HelixFormer &  CUB   & Stanford Cars                  \\ \hline
Baseline        & 0        & 66.08±0.50  & 56.55±0.52   \\
Baseline with S$\leftrightarrow$Q       & 1  & \textbf{79.34}±0.45 & \textbf{75.46}±0.37   \\
Baseline with S$\leftrightarrow$Q       & 2  & 75.03±0.49 &  74.83±0.35      \\ \hline
\end{tabular}}
\caption{The 5-way 1-shot classification results of stacking the proposed HelixFormer at the end of CNN-based backbone.}
\label{tab11}
\end{table}

We find that inserting \textit{just a single} HelixFormer at the end of CNN-based backbone is sufficiently effective in modeling the cross-image object semantic relations to boost fine-grained image classification accuracy.

To show the performance changes by stacking HelixFormer, we conduct the experiments on CUB and Stanford Cars datasets, and their results are reported in Table \ref{tab11}. It can be observed from Table \ref{tab11} that the optimal classification accuracy can be obtained when the number of HelixFormer is set to 1. This is mainly because CSRMs in HelixFormer act as a soft attention mask and filter feature representations from the CNN-based backbone network.

\setcounter{figure}{5}
\begin{figure}[!h]
\centering
\includegraphics[height=7.0cm]{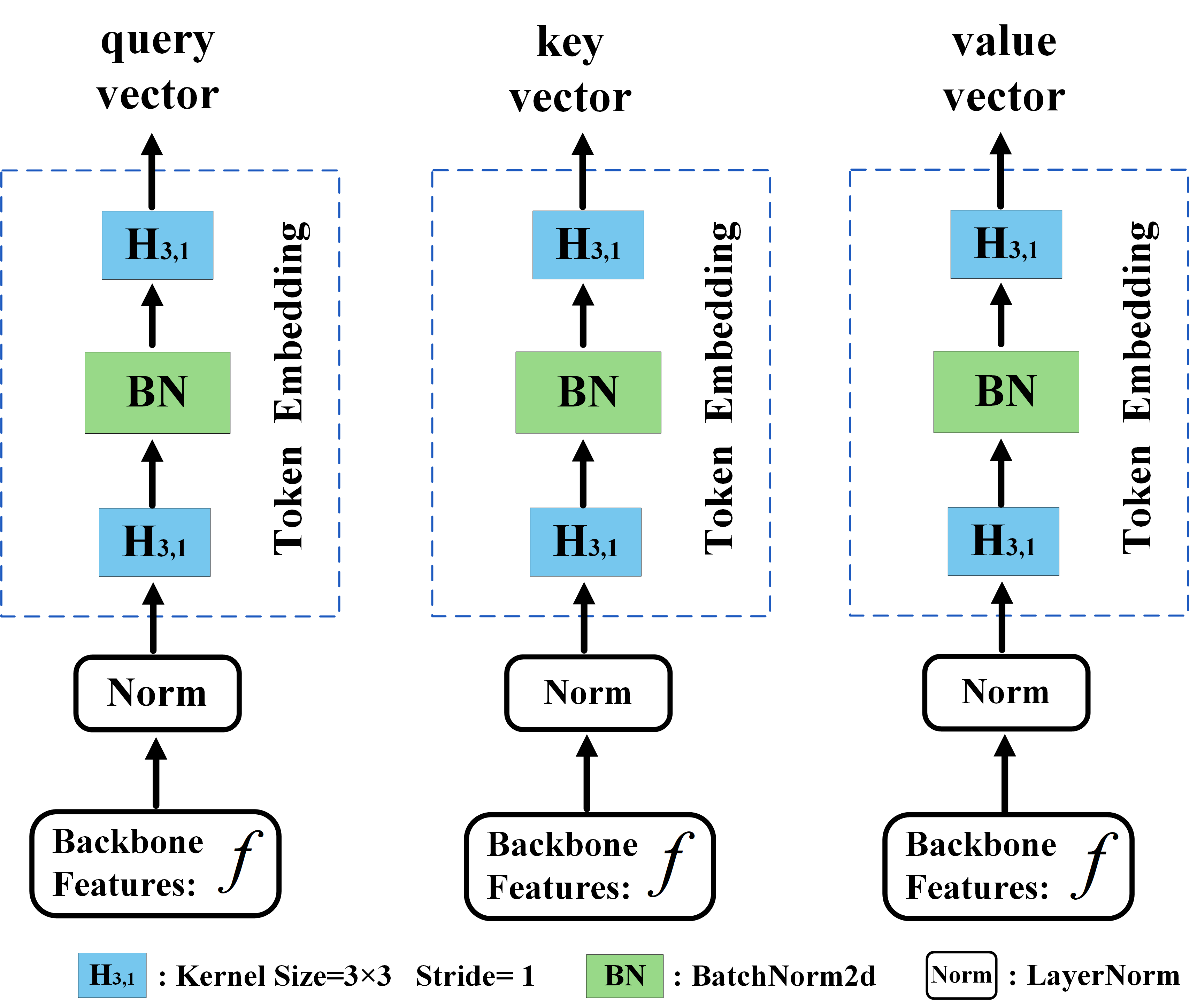}
\caption{Detailed network architecture of token embedding module in HelixFormer. All token embedding modules (including key, query, and value branches) share the same network architecture.}
\label{fig6}
\end{figure}

\setcounter{figure}{6}
\begin{figure*}[!h]
\centering
\includegraphics[height=6.1cm,width=17.2cm]{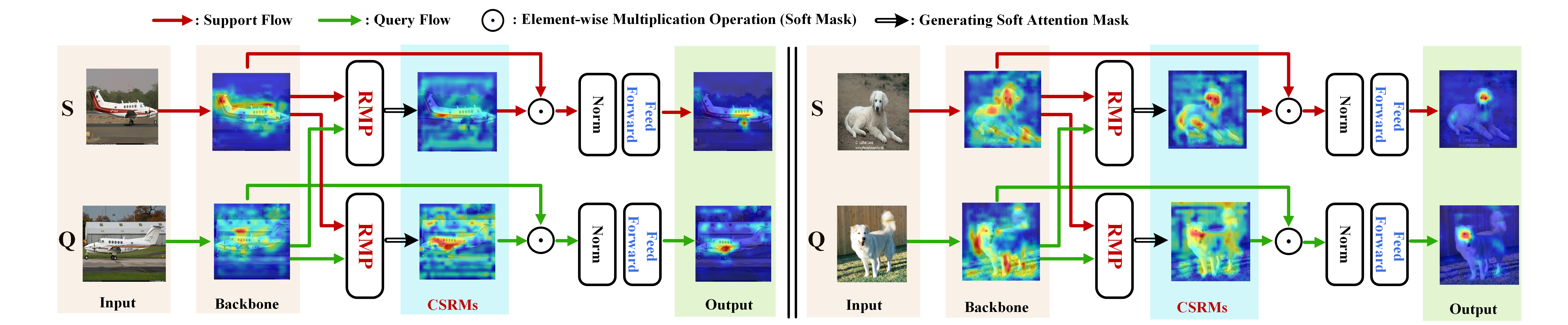}
\caption{Visualization results on Aircraft and Dogs. We visualize the features extracted by \zb{backbone, the RMP, and the REP, respectively}. Due to the local feature changes, it is hard to find the cross-image patch-level matching of semantic features, as shown in the heatmaps from the backbone. By HelixFormer, the key cross-image object semantic relations can be effectively matched.}
\label{fig7}
\end{figure*}

\setcounter{figure}{7}
\begin{figure*}[!h]
\centering
\includegraphics[height=6.1cm,width=17.2cm]{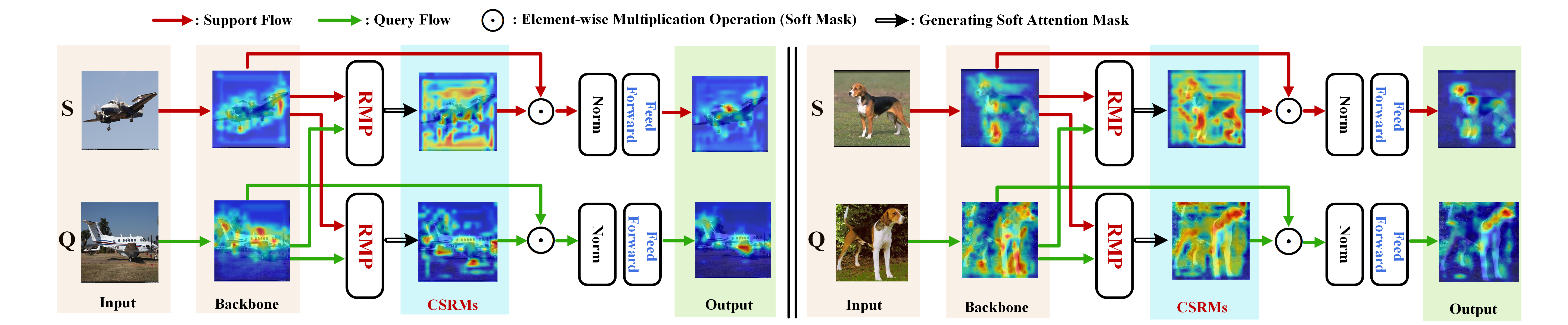}
\caption{Visualization results on Aircraft and Dogs. We visualize the features extracted by \zb{backbone, the RMP, and the REP, respectively}. Due to the local feature changes, it is hard to find the cross-image patch-level matching of semantic features, as shown in the heatmaps from the backbone. By HelixFormer, the key cross-image object semantic relations can be effectively matched.}
\label{fig8}
\end{figure*}

\setcounter{figure}{8}
\begin{figure*}
\centering
\includegraphics[height=6.1cm,width=17.2cm]{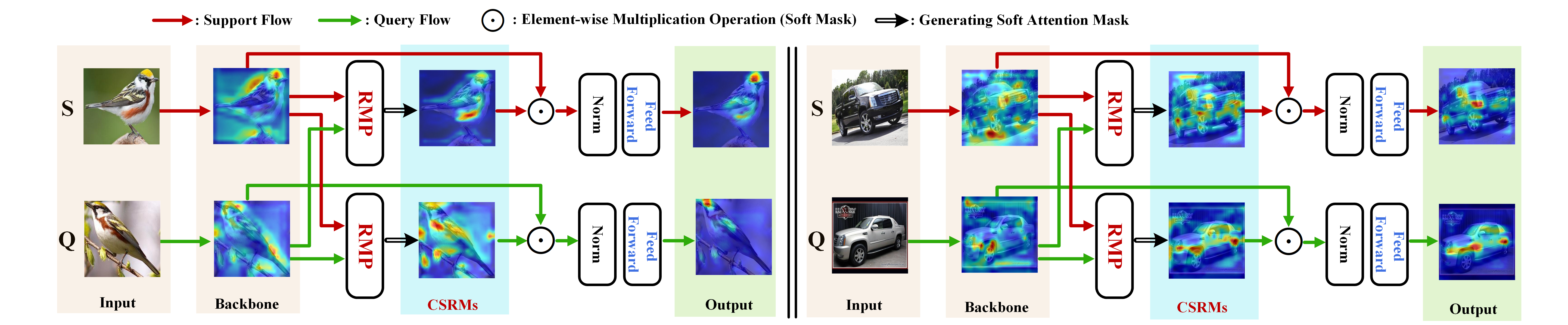}
\caption{Visualization results on CUB and Cars. We visualize the features extracted by \zb{backbone, the RMP, and the REP, respectively}. Due to the local feature changes, it is hard to find the cross-image patch-level matching of semantic features, as shown in the heatmaps from the backbone. By HelixFormer, the key cross-image object semantic relations can be effectively matched.}
\label{fig9}
\end{figure*}

\setcounter{figure}{9}
\begin{figure*}
\centering
\includegraphics[height=6.1cm,width=17.2cm]{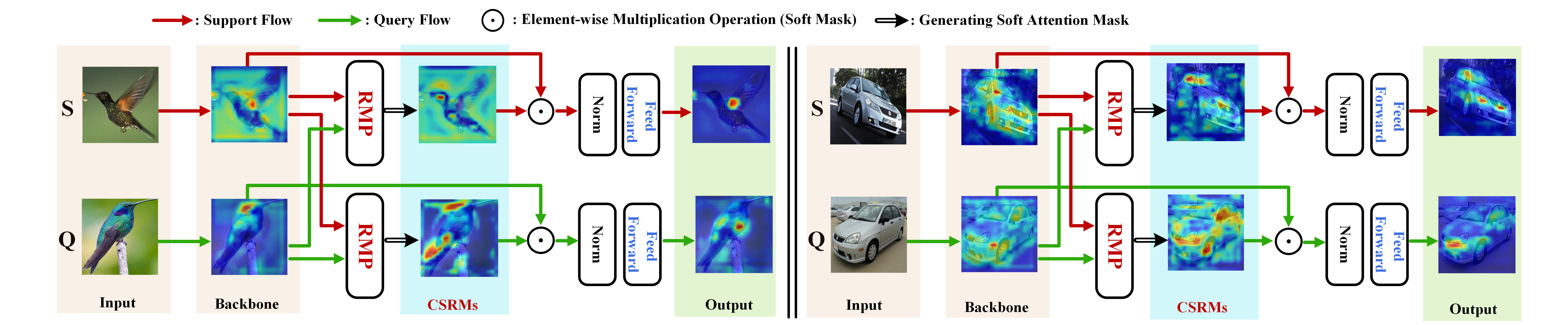}
\caption{Visualization results on CUB and Cars. We visualize the features extracted by \zb{backbone, the RMP, and the REP, respectively}. Due to the local feature changes, it is hard to find the cross-image patch-level matching of semantic features, as shown in the heatmaps from the backbone. By HelixFormer, the key cross-image object semantic relations can be effectively matched.}
\label{fig10}
\end{figure*}

\section{More Training Details and Visualization Examples}

The detailed network architecture of the token embedding module in HelixFormer is shown in Fig. \ref{fig6}, which takes the feature representations $f$ extracted by the CNN-based backbone as the input, and outputs the query, key, value vectors for the subsequent cross-attention modeling. We employ the BatchNorm layer within each token embedding module, and LayerNorm across different modules in Transformer. Besides, the standard convolution layer with a kernel size of $3\times3$ and a stride of $1$ is used as the parametric token embedding layer.

Furthermore, more visualization results of heatmaps are shown in Figs. \ref{fig7} to \ref{fig10}. These results comprehensively demonstrate that the semantically-related parts between support and query images can be found via the proposed HelixFormer.

\end{appendix}

\end{document}